\documentclass[conference]{IEEEtran}
\usepackage{graphicx}
\usepackage{balance}

\usepackage[table,xcdraw]{xcolor}
\usepackage{enumitem}

\usepackage{mathtools,amssymb,amsmath}
\usepackage{bm}
\usepackage{algorithm}
\usepackage{algpseudocode}
\usepackage{caption,subcaption}
\usepackage{url}
\usepackage{multirow}
\usepackage{changepage}
\usepackage{tikz}

\ifCLASSOPTIONcompsoc
  \usepackage[nocompress]{cite}
\else
  \usepackage{cite}
\fi

\def\BibTeX{{\rm B\kern-.05em{\sc i\kern-.025em b}\kern-.08em
    T\kern-.1667em\lower.7ex\hbox{E}\kern-.125emX}}

\newcommand\copyrighttext{%
  \footnotesize \copyright2021 IEEE. Personal use of this material is permitted. Permission from IEEE must be obtained for all other uses, in any current or future media, including reprinting/republishing this material for advertising or promotional purposes, creating new collective works, for resale or redistribution to servers or lists, or reuse of any copyrighted component of this work in other works.}
\newcommand\copyrightnotice{%
\begin{tikzpicture}[remember picture,overlay]
\node[anchor=north,yshift=-3pt] at (current page.north) {\fbox{\parbox{\dimexpr\textwidth-\fboxsep-\fboxrule\relax}{\copyrighttext}}};
\end{tikzpicture}%
}

\begin{document}

\title{Cost--effective Variational Active Entity Resolution}

\author{\IEEEauthorblockN{Alex Bogatu$^{\dag \ddag}$, Norman W. Paton$^\dag$, Mark Douthwaite$^\ddag$, Stuart Davie$^\ddag$, Andr\'e Freitas$^{\dag \nabla}$}
\IEEEauthorblockA{$^{\ddag}$\textit{Peak AI Ltd.} $^\dag$\textit{University of Manchester, UK} $^{\nabla}$\textit{Idiap Research Institute, Switzerland} \\
alex.bogatu@manchester.ac.uk}
}


% compsoc conferences compared to traditional conferences), use this
%Telephone: (800) 555--1212, Fax: (888) 555--1212}


\maketitle

\begin{abstract}
Accurately identifying different representations of the same real--world entity is an integral part of data cleaning and many methods have been proposed to accomplish it. The challenges of this entity resolution task that demand so much research attention are often rooted in the task--specificity and user--dependence of the process. Adopting deep learning techniques has the potential to lessen these challenges. In this paper, we set out to devise an entity resolution method that builds on the robustness conferred by deep autoencoders to reduce human--involvement costs. Specifically, we reduce the cost of training deep entity resolution models by performing unsupervised representation learning. This unveils a transferability property of the resulting model that can further reduce the cost of applying the approach to new datasets by means of transfer learning. Finally, we reduce the cost of labeling training data through an active learning approach that builds on the properties conferred by the use of deep autoencoders. Empirical evaluation confirms the accomplishment of our cost--reduction desideratum, while achieving comparable effectiveness with state--of--the--art alternatives.
\end{abstract}

\copyrightnotice

\section{Introduction} \label{sec:Intro}

Entity Resolution (ER), or the process of identifying different representations of the same real--world entity, has been the subject of more than 70 years of research. Yet, practitioners often opt for \textit{ad hoc} ER approaches mainly due to deficiencies in adapting existing solutions to new and specialized datasets. In practice, adapting an ER solution to a new use--case often translates to (1) mapping the types of features that govern the comparison of two entities and are expected by the adopted solution to the case at hand, i.e., \textit{feature engineering}; (2) identifying local examples of duplicates and non--duplicates, i.e., \textit{data labeling}; and (3) learning a task--specific similarity function that discriminates between duplicates and non--duplicates, i.e., \textit{similarity learning}. Despite the sheer number of ER methods available \cite{elmagarmid-2007}, few solutions manage to reduce the costs incurred by the above triple, costs often paid in active user involvement and long configuration and training times. 

With respect to (1) and (3) above, recent proposals (e.g., \cite{ebraheem-2018, mudgal-2018}) resort to deep learning techniques \cite{goodfellow-2016} to adapt general--purpose features (e.g., word--embeddings \cite{bengio-2000}) to new ER use--cases and learn a task--specific similarity function \textit{concomitantly}. However, this is done at the expense of (2), since the resulting solutions tend to have a voracious appetite for labeled data (e.g., up to thousands of labeled instances \cite{mudgal-2018}), and at the expense of training time, since such approaches can take up to hours to generalize. With respect to (2) above, recent deep ER--focused active learning proposals (e.g., \cite{kasai-2019}) aim to generate labeled data for merely adapting an existing and already trained entity matching model to the case at hand.
% assume the existence of an already trained matching model aim to generate reduced amounts of labeled samples, intended to merely improve an already optimized transferred model. 
Therefore, there is limited support for manually labeling the volumes of training data often required by deep learning ER.

In this paper, we aim to reduce the costs associated with performing deep learning--based ER in practice by pioneering the use of Variational Auto-Encoders (VAEs) \cite{kingma-2014} for automatically generating entity representations. This allows us to \textit{decouple feature engineering from similarity learning}, perform the former without user supervision (step \textit{1} in Figure \ref{fig:er}), and only rely on supervised learning for the latter (step \textit{2} in Figure \ref{fig:er}). 
% Additionally, the decoupling conveys a \textit{transferability} property to the resulting feature learning model, i.e., use it across different ER tasks, and an \textit{adaptability} property to the similarity learning model, i.e., adapt the feature space to the case--at--hand through existing labeled data. 
Additionally, we support the data labeling effort for the supervised step through a proposed active learning technique (step \textit{3} in Figure \ref{fig:er}), facilitated by the above--mentioned decoupling. Therefore, our central contribution in this paper is an ER method that learns similarity functions over an unsupervised feature space and we show its cost--effectiveness potential through:

\begin{figure}[t]
\centering
\captionsetup{justification=centering}
\includegraphics[width=.7\linewidth]{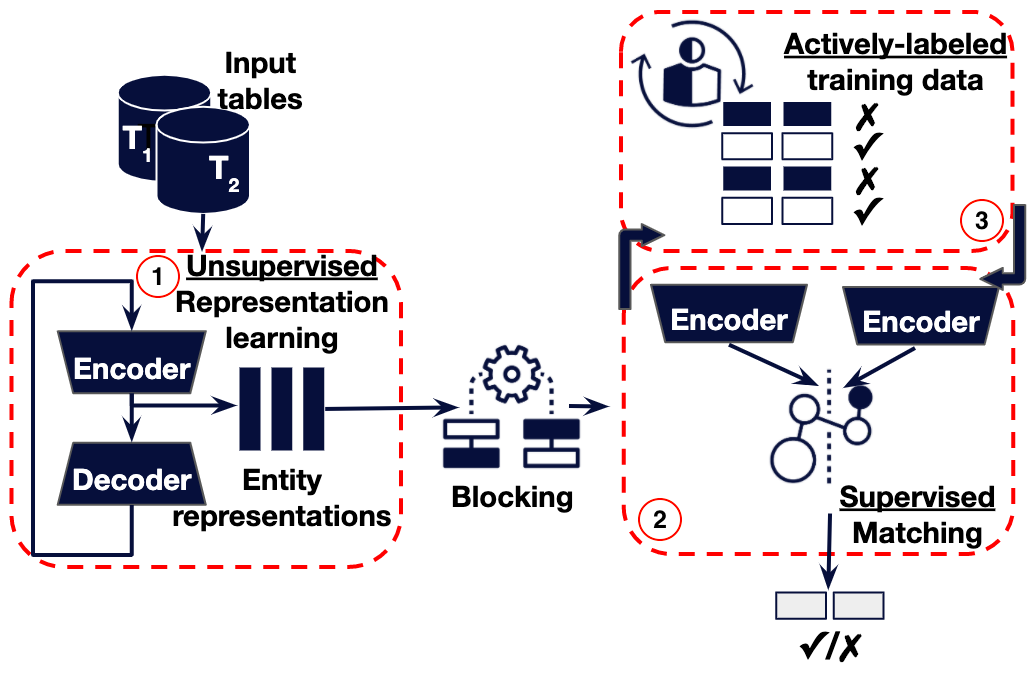}
\caption{\small Decoupled, cost--effective ER process}
\label{fig:er}
\end{figure}

\begin{itemize}[leftmargin=*, nosep]
    \item An \textit{unsupervised} and \textit{transferable} representation learning method for producing similarity--preserving feature vectors for tuples (i.e., entities). Unsupervised because it builds on deep generative models (e.g., VAE) to automatically generate the feature vectors. Transferable because the resulting model is reusable without re--training across different ER scenarios and data domains.
    \item An \textit{adaptable} matching method for learning task--specific similarity functions. Adaptable because it builds on Siamese neural networks \cite{bromley-1993} to fine--tune previously learned tuple representations to better reflect the notion of similarity derived from given training data.
    \item An \textit{active--learning} scheme for assisting the user in labeling supervised matching training instances. Our contribution here is the exploitation of the generative property of VAEs to identify tuple pairs that are informative, diverse, and balanced with respect to the match/non--match classes.
    \item An empirical evaluation of the above on a multitude of domains, demonstrating their collective potential for ER cost reduction in terms of (i) feature engineering and similarity learning; (ii) training times; and (iii) data labeling efforts.
\end{itemize}

\section{Background and related work} \label{sec:Relwork}

Typically, ER methods for structured data (e.g., \cite{elmagarmid-2007}) include at least a \textit{blocking} and a \textit{matching} step \cite{maskat-2016}. 
% The purpose of the former is to efficiently identify pairs of entities for which there is some evidence of similarity and limit running the matching step on those candidates only \cite{papadakis-2016}. 
The latter, which is the focus of this paper, involves detailed comparison of candidates efficiently identified by the former, often resorting to rule--based reasoning (e.g., \cite{fan-2009, singh-2017}), classification (e.g., \cite{bilenko-2003, konda-2016, ebraheem-2018}), crowdsourcing (e.g., \cite{wang-2012}), or generative modeling (e.g., \cite{wu-2020}) to do so.

% However, we note that performing blocking on the tuple representations generated by our representation learning module is straightforward using approximate nearest--neighbour algorithms (e.g., \cite{indyk-1998, ram-2019}), similar to \cite{ebraheem-2018} and as shown in Section \ref{sec:Evaluation}.

In this paper, we focus on \textit{classification--based} matching. In practice, it often relies on string similarities between attribute values of tuples (e.g., \cite{bilenko-2003, konda-2016}). Alternatively, NLP--specific feature engineering methods (e.g., \cite{mikolov-2013, goodfellow-2016}) are employed to extract meaningful numerical features from attribute values. These are then passed to deep learning \cite{goodfellow-2016} classifiers that learn a similarity function consistent with given duplicate/non--duplicate training examples \cite{mudgal-2018, ebraheem-2018, nie-2019}. Our proposal in this paper, \textbf{V}ariational \textbf{A}ctive \textbf{E}ntity \textbf{R}esolution (\textit{VAER}), is a \textit{deep learning ER} solution that focuses on reducing the human--involvement cost of performing ER in practice. Specifically, existing supervised ER proposals, such as \textit{DeepER} \cite{ebraheem-2018}, \textit{DeepMatcher} \cite{mudgal-2018}, \textit{Seq2SeqMatcher} \cite{nie-2019}, or \textit{DITTO} \cite{yuliang-20} aim to perform feature engineering and match training simultaneously, using the same training instances. This leads to complex models that require thousands of labeled instances, potentially hours of training \cite{mudgal-2018}, and that are not reusable. Similarly, unsupervised approaches, such as \textit{ZeroER} \cite{wu-2020}, involve costly feature engineering efforts for every new ER task. Conversely, \textit{VAER} decouples the feature engineering and matching tasks, conducting the former in an unsupervised fashion and only optimizing the latter on account of training data. Crucially, this leads to significantly smaller training times for matching and, consequently, enables the use of the resulting model in iterative active learning strategies that seek to assist the user in generating training data. Moreover, the feature engineering task in \textit{VAER} builds on a \textit{representation learning} paradigm \cite{bengio-2013} that enables the transfer of the resulting representation model to multiple different ER tasks, i.e. transfer learning \cite{pan-2010}.

In supporting the above characteristics, the techniques used with \textit{VAER} intersect with the following research areas:
\smallbreak

\noindent\textbf{Variational Auto--Encoders}. VAEs \cite{kingma-2014, rezende-2014} are a type of neural network often used in dimensionality reduction, representation learning and generative learning. Typically, a VAE involves an \textit{encoder} and a \textit{decoder} that are trained simultaneously. The encoder aims to approximate a (lower dimension) probability distribution of the input using variational inference \cite{jordan-1999, beal-2003}. The decoder assists the encoder by ensuring that any random sample from the approximated distribution can be used to reconstruct the input. Therefore, VAEs can be seen as unsupervised models, since the labels are the inputs themselves. In this paper, we use a VAE to perform unsupervised entity representation learning. Formal details about our VAE--based approach are further provided in Section \ref{sec:RL}.
\smallbreak

\noindent\textbf{Active Learning}. AL is a sub--field of machine learning based on the key hypothesis that if a learning algorithm can suggest the data it learns from, it could perform better with less training \cite{burr-2009}. In ER, AL has been used to ease the user's task of labeling training data (e.g., \cite{qian-2017, kasai-2019, meduri-2020}. For example, \cite{meduri-2020} offers a framework for AL in ER with various types of non--deep learning algorithms. With respect to deep learning ER, in \cite{kasai-2019}, new training samples are used to adapt a transferred matching model from another ER use--case to the case at hand. While our approach follows a similar methodology, we do not assume pre--trained matching models and start from very few labeled instances to create an, often weak, initial model that is then iteratively improved based on evidence specific to and dictated by the nature of our entity representations.

\section{Unsupervised representation learning} \label{sec:RL}

With respect to ER feature engineering, we need to convert each input tuple into a numeric vectorized representation expected by downstream matching tasks. The viability of such representations ultimately determines the effectiveness of the entire ER process. Viable representations compact most of the high--level similarity--salient factors into dense vectors that are close together in their multivariate space for duplicates and far apart for non--duplicates. In this section, we set out to generate such representations, albeit in the more expressive form of probability distributions, rather than fixed vectors. Here, we emphasize a \textit{first cost--effectiveness property} of our system: we generate tuple representations unconstrained by the need for training data or user decisions with respect to data characteristics that govern the similarity of two entities.

\subsection{Entity representation architecture} \label{subsec:Arch}

The intuition in this section is that the attribute values of duplicate tuples originate from similar prior distributions that encode the information conveyed by the values. While there is no constraint on the input data distribution itself, we attempt to approximate the prior distributions as Gaussians by using a VAE with \textit{shared parameters} across attributes\footnote{{\small By ``shared" we mean that the model will generate representations for all attribute values of a tuple simultaneously by processing a 2--d input: \textit{num. attributes} $\times$ \textit{num. features}}}. This distribution type is a constraint on the VAE model, not the input data, and is dictated by the need for analytical interpretation and stability of the training process. One other distribution with similar properties is the von Mises--Fisher \cite{davidson-18}. A sufficiently powerful VAE with non--linearity can map arbitrary distributions to the Gaussian/von Mises--Fisher and back, so from a theoretical viewpoint, the only constraint for the choice of latent distribution is given by the VAE training process, i.e, the smoothness of the latent space \cite{kingma-2014}.

\begin{figure}[t]
\centering
\includegraphics[width=.75\linewidth]{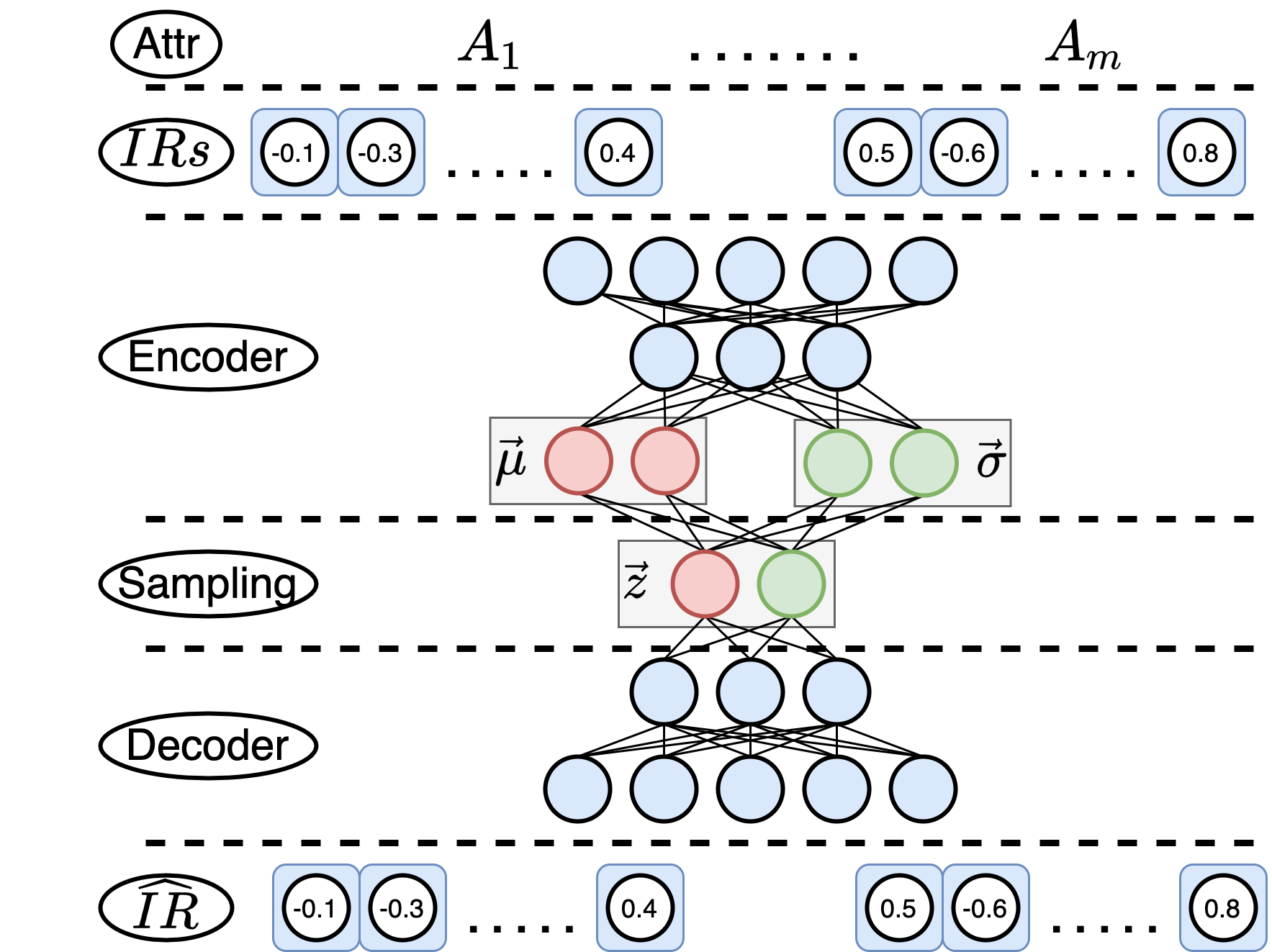}
\caption{\small Proposed entity representation model architecture}
\label{fig:arch}
\end{figure}

Broadly, given a tuple with $m$ attribute values denoted $\{A_1, \ldots A_m\}$, for each $A_i$ we want to approximate a distribution over some random variable which encodes both \textit{morphological} (i.e., syntactic form of words) and \textit{semantic} (i.e., natural language meaning) factors. The first step towards this goal is to map individual attribute values to dense vectors, which we call \textit{Intermediate Representations} ($IRs$), capable of capturing the similarity between close attribute values. Then, we proceed to approximating a distribution over $IRs$ that will allow us to probabilistically reason about their similarity. Assuming for now the existence of $IRs$, Figure \ref{fig:arch} illustrates the overall neural architecture of the entity representation model proposed in this paper.

\noindent\textbf{Attr}. The attribute values layer where we treat each attribute value independently. Consequently, tuple comparison at matching time can be more granular, i.e., comparing attribute representations pair--wise. Furthermore, attribute--level weighted matching schemes can also be employed.
\smallbreak

\noindent$\boldsymbol{IR}$. The Intermediate Representation ($IR$) layer where attribute values are transformed into initial vectorized representations that encode semantic and morphological factors. These representations are the inputs to the VAE and are further described in the next subsection.
\smallbreak

\noindent\textbf{Encoder}. The encoding layer that takes a collection of $IRs$ as input, passes them through one or more \textit{dense} neural layers with \textit{non--linear} activation functions, e.g., rectified linear functions (ReLU), and approximates a latent multivariate Gaussian distribution with diagonal covariance, $\mathcal{N}(\mu, \sigma)$, for each $IR$. The mean and covariance diagonal parameters of distributions produced by the \textit{Encoder}, $\{(\mu_1, \sigma_1), \dots$ $(\mu_m, \sigma_m)\}$, one for each attribute value, denote the desired entity representations used in downstream tasks to identify duplicates. In other words, each tuple will be represented by a collection of $(\mu, \sigma)$ pairs and the comparison of two tuples will be performed attribute--wise by comparing the corresponding distributions.
\smallbreak

\noindent\textbf{Sampling}. This layer performs ancestral sampling from each $\mathcal{N}(\mu, \sigma)$ following a procedure specific to VAEs known as the \textit{reparameterization trick} \cite{kingma-2014}. This allows training a VAE by means of backpropagation with (stochastic) gradient descent \cite{kushner-2003} that requires the model to be deterministic. Moreover, this step confers a generative property to the representation model: given an attribute value $v$, each sample $z$ generated from the corresponding $\mathcal{N}(\mu_v, \sigma_v)$ can faithfully encode the latent characteristics of $v$. 
% We further exploit this in Section \ref{sec:AL}.
\smallbreak

\noindent\textbf{Decoder}. The decoding layer that reverses the encoding to reconstruct the original input (i.e., $IRs$), conditioned by latent variables $z$ randomly sampled from $\mathcal{N}(\mu, \sigma)$.
\smallbreak

\noindent$\boldsymbol{\widehat{IR}}$. The last layer of the architecture denotes the reconstructed $IRs$ that, ideally, are as close as possible to the original $IRs$. The difference between the input $IRs$ and the output $\widehat{IRs}$ represents one of the minimizing objectives used in training the model, as described in Section \ref{subsec:learning}.

Given a collection of $IRs$, the components described above are trained in tandem. Before describing this training process we first discuss how we obtain $IRs$ and their importance.

\subsection{Intermediate representations of attributes} \label{subsec:IR}

Deep--learning models, including VAEs, operate on numerical inputs, i.e., feature vectors. One possible approach to obtain such representations is, similarly to \cite{ebraheem-2018}, \cite{mudgal-2018} or \cite{yuliang-20}, to use a RNN-- \cite{ebraheem-2018} or BERT--based neural architecture \cite{devlin-19}. However, such an approach would require considerable training resources and, more importantly, would bound the resulting representations of tuples to the current data domain and ER task, since they rely on processing the word vocabulary of all input tuples. The resulting representations would then be hard to reuse in other ER tasks.

In this paper, we rely on a simple alternative that generates initial attribute value vectorized representations, i.e., \textit{intermediate representations ($IRs$)} that are \textit{similarity--preserving} and independent from the type of neural architecture used for matching. Such $IRs$ are vectors of numbers that can be independently generated using methods such as Latent Semantic Analysis (\textit{LSA}) \cite{dumais-2004}, word--embedding models \cite{pennington-2014, mikolov-2013} (\textit{W2V}), BERT pre--trained embeddings \cite{Wolf-20}, or relational embeddings specialized for data integration tasks such as ER \cite{cappuzzo-20} (\textit{EmbDI}). In practice, $IR$ generation involves construing each attribute value of a given table as a sentence and:
\begin{itemize}[leftmargin=*, nosep]
\item \textit{LSA}: consider the corpus of all such sentences to model its latent topics and generate $IRs$ using known topic modeling methods %such as Latent Semantic Analysis (LSA) 
(e.g., \cite{dumais-2004,blei-2003}). % or Latent Dirichlet Allocation \cite{blei-2003}.
\item \textit{W2V}: pass each word of each sentence through a pre--trained word--embedding model and generate $IRs$ by averaging the resulting embeddings at sentence--level.
\item \textit{BERT:} pass each sentence through a pre--trained BERT model \cite{Wolf-20} and construe the result as the $IR$ for the input sentence (i.e., attribute value).
\item \textit{EmbDI:} consider the corpus of all sentences to train a data integration embedding model following \cite{cappuzzo-20} and follow the \textit{W2V} method to obtain $IRs$.
\end{itemize}

$IRs$ are important because they encode morphological and semantic information of values. Without their use, we would have to rely on the VAE architecture to learn numerical representations of words using high--dimensional deep \textit{Embedding} layers. Such architecture types are common with \textit{DeepMatcher} or \textit{DITTO}. In this paper, we show that \textit{VAER} is general enough so that it can incorporate the types of evidence commonly used with other ER systems (e.g., \textit{W2V} in \textit{DeepMatcher} and \textit{BERT} in \textit{DITTO}), while further contributing to the cost reductions of the ER process by (i) reducing the dimensionality of the VAE; (ii) speeding up the VAE training process, e.g., through transfer learning, as described in Section \ref{subsec:transf}; and (iii) reducing the generalization requirements, since part of the similarity correlations between words aimed to be conveyed by the final representations are already caught by $IRs$.

\subsection{Learning entity representations: training the VAE} \label{subsec:learning}

$IRs$ can themselves act as entity representations. However, they are deterministic, fixed vectors in a potentially irregular latent space with reduced control over how close duplicates end up. In practice, $IRs$ tend to be effective for clean and structured data and less so when data is dirty. The model from Figure \ref{fig:arch} attempts to address such limitations by learning a probabilistic model over $IRs$. Specifically, it encodes a given input, $IR$ in our case, as a probabilistic latent variable $z$ that captures the high--level information of the input, and then decodes that input (or a close version of it) from $z$. The objective of the encoding--decoding process is to minimize the error between the input and the output \cite{kingma-2014}.

More formally, given a collection of $n$ entities with $m$ attributes and each entity represented by the $m$ $IRs$ of its attribute values, $\{\{IR^1_1, \ldots IR^1_m\}, \ldots$ $\{IR^n_1, \ldots IR^n_m\}\}$, we consider each $IR$ to be a random variable generated by some random process involving a lower--dimension latent variable $z$ drawn from some prior distribution $p(z)$ that conveys high--level similarity--salient factors of $IR$. We want to infer the characteristics of $z$, given $IR$. In other words, we want to compute a posterior distribution $p(z|IR)$, given by $p(z|IR) = \frac{p(IR|z)p(z)}{p(IR)}$. Computing the denominator $p(IR) = \int p(IR|z)p(z)dz$ is intractable due to the multidimensionality of $IR$ and $z$ \cite{kingma-2014}. Alternatively, $p(z|IR)$ can be approximated through \textit{variational inference} \cite{beal-2003} by another tractable (e.g., Gaussian) distribution $q(z|IR)$ \cite{kingma-2014}. In practice, this translates to minimizing the Kullback–Leibler (KL) divergence between $q(z|IR)$ and $p(z|IR)$ which can be achieved by maximizing:

\begin{equation} \label{eq:obj}
    \small
    \mathbb{E}_{q(z|IR)}log(p(IR|z)) - KL(q(z|IR)||p(z))
\end{equation}

\noindent where the first term represents the \textit{expected log--likelihood}) of faithfully reconstructing $IR$ given some $z$ from $q(z|IR)$, and the second term is the KL divergence between our approximated distribution $q(z|IR)$ and the true prior. Consistently with our initial assumption, Equation \ref{eq:obj} ensures that $q(z|IR)$ describes a distribution of faithful latent representations of $IR$, so that $IR$ can be reconstructed from its samples, and that any latent representation $z$ comes from a similar distribution to the assumed prior $p(z)$. In practice, by fixing $p(z) = \mathcal{N}(0, I)$ (i.e., the standard normal distribution of mean $0$ and diagonal unit covariance), the VAE enforces a regular geometry on the latent space so that similar data lead to similar latent variables \cite{bowman-2016}.

From a practical perspective, we can construct the inference model described above into a neural network model where a function $\phi: \mathbb{R}^d \rightarrow \mathbb{R}^k$, i.e., our \textit{Encoder} from Figure \ref{fig:arch}, maps a $d$--dimensional input, $IR$, to two $k$--dimensional variables, $\mu$ and $\sigma$, denoting the parameters of a latent Gaussian distribution, $q_{\phi}(z|IR)$. Additionally, a second function, $\theta: \mathbb{R}^k \rightarrow \mathbb{R}^d$, i.e., our \textit{Decoder} from Figure \ref{fig:arch}, ensures that any latent variable $z$ sampled from $q_{\phi}(z|IR)$ can be used to produce an approximate reconstruction of $IR$, (i.e., $\widehat{IR}$). The loss function minimized in learning the parameters of $\phi$ and $\theta$ follows the relation from Equation \ref{eq:obj}, extended to include all $m$ $IRs$ (i.e., attribute values) of a tuple:

\begin{equation} \label{eq:real_obj}
\small
\begin{split}
   L_{(\phi, \theta)}(IR, \widehat{IR}) & = \sum_{i=1}^m \underset{q_{\phi}(z_i|IR_i)}{\mathbb{E}}[log(p_{\theta}(IR_i|z_i))] \\
    & - \sum_{i=1}^m KL(q_{\phi}(z_i|IR_i)||\mathcal{N}(0, I))
\end{split}
\end{equation}

Intuitively, by fitting the inputs to Gaussian distributions, the VAE learns representations of attribute values not as single points, but as ellipsoidal regions in the latent space, forcing the representations to continuously fill this space. Specifically, the mean $\mu$ of the latent distribution controls where the encoding of an input should be centered around, while the diagonal covariance $\sigma$ controls how far from the center the encoding can vary. As decoder inputs are generated at random from anywhere inside this distribution (recall the \textit{Sampling} layer of Figure \ref{fig:arch}), the decoder is exposed to a range of variations of the encoding of the same input during training. The decoder, therefore, learns that, not only is a single point in the latent space referring to an attribute value, but all nearby points refer to that value as well, i.e., accounting for variation and uncertainty across the attribute values of duplicates.

% Lastly, by learning independent representations for each attribute value, we obtain \textit{disentangled} representations of tuples, $\{(\mu_1, \sigma_1), \dots (\mu_m, \sigma_m)\}$. By disentangled, we mean a representation where single latent units (e.g., $(\mu_i, \sigma_i)$) are sensitive to changes in single attribute values, while being relatively invariant to changes in other attribute values.

\subsection{Representation model transferability} \label{subsec:transf}

The architecture from Figure \ref{fig:arch} enables a \textit{second cost--reduction characteristic} to our overall approach: the representation model trained during one ER task can be reused in other ER tasks, therefore eliminating the need for representation learning. In other words, the architecture from Figure \ref{fig:arch} allows for transfer learning \cite{pan-2010}. This is by virtue of the variational inference and the use of $IRs$ as inputs. Concretely, because the model from Figure \ref{fig:arch} operates on numerical $IRs$, its output distributions are \textit{independent from domain--specific aspects}, such as the domain vocabularies. Consequently, once the parameters of the VAE are trained, any new $IR$ with \textit{similar dimensionality} to the one required by the transferred model architecture can be accurately encoded, regardless of the $IR$'s data domain. However, these new $IRs$ have to already convey similarity signals since the pre--trained $VAE$ can only amplify existing ones. In practice, using $IRs$ of the types discussed in Section \ref{subsec:IR} satisfies this requirement. It follows that the transferability property is \textit{domain--agnostic} and can eliminate the need for feature engineering from new ER tasks while minimizing the training--time costs, since representation learning accounts for most of the training time needs.

\section{Supervised matching in the latent space} \label{sec:Matching}

The transferability property is complemented by an adaptability property. Specifically, the representations produced by the model from Figure \ref{fig:arch} are lenient with respect to small variations in the attribute values of two duplicate tuples. However, more significant discrepancies between duplicates that are not reflected in the $IRs$ can lead to far apart latent distributions, especially if the representation model has been transferred from another use--case. Moreover, since the representation learning step is unsupervised, the notion of similarity between tuples conveyed by the learned representations may not be consistent with the real intended notion. For instance, consider the example from Table \ref{tab:dupex}. Both entities denote the same song by the same artist released as part of two different albums. Whether or not the two tuples are duplicates depends on the use--case and the unsupervised representations may not reflect that decision. There is, therefore, a need for (i) adjusting the entity representations to cover more significant discrepancies between duplicates, and (ii) aligning the notion of similarity conveyed by the representations with the use--case intent, i.e, similarity learning. In this section, we describe our supervised deep learning proposal for addressing these requirements.

\begin{table}[t]
     \centering
     \footnotesize
     \caption{\small Duplicate candidates songs example}
     \begin{tabular}{|c|c|c|c|}
     \hline
     Song & Artist & Album & Year \\  
     \hline
     \hline
     Charlie Brown & Coldplay & Mylo Xyloto & 2011 \\
     \hline
     Charlie Brown & Coldplay & GRAMMY Nominees & 2013 \\
     \hline
     \end{tabular}
     \normalsize
     \label{tab:dupex}
\end{table}

\subsection{Matching architecture} \label{subsec:m_arch}

\begin{figure}[t]
\centering
\includegraphics[width=.7\linewidth]{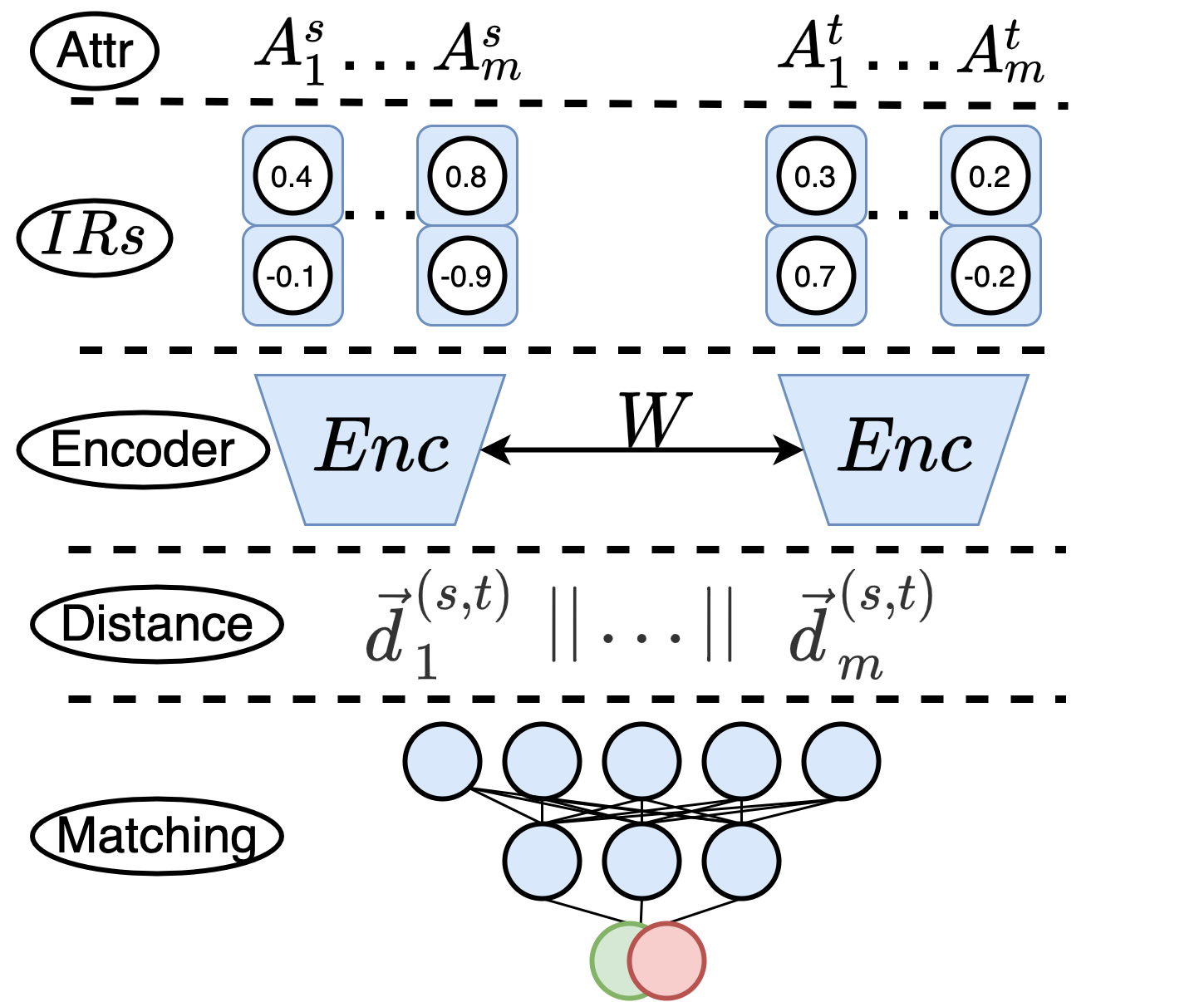}
\caption{\small Proposed matching model architecture}
\label{fig:m_arch}
\end{figure}

In Section \ref{sec:RL} we focused on a \textit{generative} task of producing similarity--preserving entity representations. We now describe a \textit{discriminative} task that builds on the generative model to learn a similarity measure between representations and to perform ER matching. More specifically, given a set of tuple pairs $(s, t)$, each with $m$ attribute values $\{A^s_1, \ldots A^s_m\}$ and $\{A^t_1, \ldots A^t_m\}$, and corresponding duplicate/non--duplicate labels, we perform supervised training of a Siamese neural network (i.e., a class of neural network that contain two or more identical sub--networks) \cite{bromley-1993, neculoiu-2016}. Our proposed architecture is illustrated in Figure \ref{fig:m_arch}.

\noindent\textbf{Attr} and $\boldsymbol{IR}$. These layers correspond to the first two layers from Figure \ref{fig:arch}: each attribute value of the input tuples is mapped to an $IR$ which is then passed to the next layer.
\smallbreak

\noindent\textbf{Encoder}. This layer uses two variational encoders, similar to the one from Figure \ref{fig:arch}. Both encoders share the same weights initialized with the trained values of the variational encoder in Figure \ref{fig:arch}. Parameter updating is mirrored across both encoders during training and each encoder generates an entity representation, $\{(\mu_1, \sigma_1), \dots (\mu_m, \sigma_m)\}$, corresponding to its input tuple. The purpose of this layer is to improve the weights transferred from the variational encoder of the representation model on account of training data.
\smallbreak

\noindent\textbf{Distance}. Modeling the latent space using probability distributions allows us to reason about the similarity of $s$ and $t$ in terms of the distance between their corresponding distributions. Two examples of metrics that can be used to quantify such distance between Gaussian distributions are Wasserstein \cite{mallasto-2017} and Mahalanobis \cite{gallego-2013}. These have previously been used in learning sentence similarity in NLP (e.g., \cite{deudon-2018}). In practice, both distances showed similar effectiveness, so we are only discussing the former here.

Intuitively, the $d$--Wasserstein distance quantifies the minimal cost of transporting the unit mass of one probability measure into the unit mass of another probability measure, when the cost is given by an $L^d$ distance \cite{mallasto-2017}. In our case, we consider $d=2$ and the squared $2$--Wasserstein distance ($W^2_2$) between two $k$--dimensional diagonal Gaussian distributions, $p$ and $q$, is given by Equation \ref{eq:wasserstein}.

\begin{equation} \label{eq:wasserstein}
\small
    W^2_2(p, q) = \sum^{k}_{i=1} (\mu^p_i - \mu^q_i)^2 + (\sigma^p_i - \sigma^q_i)^2
\end{equation}

Returning to the \textit{Distance} layer in Figure \ref{fig:m_arch}, from the two inputs $\{(\mu^s_1, \sigma^s_1), \dots (\mu^s_m, \sigma^s_m)\}$ and $\{(\mu^t_1, \sigma^t_1), \dots (\mu^t_m, \sigma^t_m)\}$, we compute $m$ attribute--wise Wasserstein distance vectors $\vec{d}^{(s,t)} = (\mu^s - \mu^t)^2 + (\sigma^s - \sigma^t)^2$. Then, we concatenate all $m$ distance vectors and pass the result to the next layer. 

\smallbreak

\noindent\textbf{Matching}. This layer performs ER matching in the form of a binary classification task. The classifier consists of a two--layer Multi Layer Perceptron (MLP) with non--linear activation functions that, given the $m$ concatenated Wasserstein distance vectors, predicts a match/non--match label. The purpose of this layer is to discriminate between duplicates and non--duplicates.

% Given training data, the model from Figure \ref{fig:m_arch} undergoes a training process based on backpropagation and (stochastic) gradient decent, with the weights of \textit{Matching} and \textit{Encoder} layers being updated at each step, as we now describe. 

In practice, at inference time when there may be many tuple pairs to match, although the distance computation and matching overheads are neglectable compared to the training--time costs of \textit{VAER}, they could be alleviated by a blocking strategy (e.g., as in  \cite{ebraheem-2018}) that filters out clear non--duplicates. We describe how such a step could be adapted to our distribution--based representations in Section \ref{subsec:rl_exp}.

\subsection{Learning entity similarity: training the Siamese network}

The training process of the proposed matching model involves two optimization objectives: (1) minimize the $W^2_2$ distance between the representations of duplicates and maximize it for non--duplicates, i.e., improving the \textit{Encoder} layer, and (2) minimize the classification error of the binary classifier, i.e., training the \textit{Matching} layer. These two objectives are optimized simultaneously. Firstly, with respect to (1), we further improve the initial weights of the Siamese encoder heads on account of training data so that they are consistent with more complex cases of tuple similarity, infeasible to cover by the unsupervised approach alone. Secondly, in relation to (2), we optimize the parameters of the binary classifier such that the resulting model is effective in discriminating between duplicate and non--duplicate tuples. To this end, we use a contrastive loss function \cite{neculoiu-2016} defined by Equation \ref{eq:contrastive}:

\begin{equation} \label{eq:contrastive}
\small
\begin{split}
    L(s, t) & = ylog(p_{\gamma}(y|d^{(s,t)})) + (1-y)log(1-p_{\gamma}(y|d^{(s,t)})) \\
    & + \frac{1}{m}\sum_{i=1}^m [xW^2_2(q_{\phi}(A^s_i), q_{\phi}(A^t_i)) \\
    & + (1-x)\max (0, M - W^2_2(q_{\phi}(A^s_i), q_{\phi}(A^t_i)))]
\end{split}
\end{equation}

\noindent where $y$ is the predicted class for a pair of tuples $(s, t)$, $x$ is the true class, $p_{\gamma}(y)$ is the predicted probability of $y$, $\gamma$ are the parameters of the binary classifier, $d^{(s,t)}$ is the distance vector from the \textit{Distance} layer in Figure \ref{fig:m_arch}, $q_{\phi}(A^j_i)$ is the Gaussian distribution approximated for attribute $A^j_i$ by the encoding layer, and $\phi$ are the parameters of the encoders. $M$ is a margin hyperparameter that controls how the encoders weights are adjusted in light of the training data.

The first term in Equation \ref{eq:contrastive}, i.e., the cross--entropy of the prediction, covers objective (2) from above, while the second term covers objective (1). The function of the margin $M$ is that, when the representations produced for a negative pair are distant enough, no effort is wasted on maximizing that distance, so further training can focus on more difficult pairs.

The cost benefits of the feature and similarity learning approaches described in Sections \ref{sec:RL} and \ref{sec:Matching} are now clear: given two input tables (as in Figure \ref{fig:er}), the unsupervised representation learning model from Figure \ref{fig:arch} alleviates the user from deciding on the features that govern the tuple comparison process, and from providing training data to generate those features. The potential loss in feature expressiveness determined by the unsupervised nature of the representation model, or by the use of a transferred representation model, is attenuated by the Siamese matching model from Figure \ref{fig:m_arch} that adjusts the parameters of the representation encoder in light of training data. This adjustment is also efficient, since most of the optimization work has already been done by the unsupervised training process.

\section{Active learning in the latent space} \label{sec:AL}

Deep learning solutions are often characterized by the need for a significant number of training instances (e.g., up to thousands \cite{mudgal-2018}). Active learning (AL) \cite{burr-2009} has traditionally been proposed to support the manual labeling of such volumes where an initial pool of labeled instances, $\mathcal{L}$, and a much larger pool of unlabeled instances, $\mathcal{U}$, are assumed. Then, the user is tasked with labeling one or more instances from $\mathcal{U}$ to iteratively update $\mathcal{L}$. The crux of an AL strategy is the sampling method used to choose instances from $\mathcal{U}$ to be passed to the user for labeling. Many such strategies proposed for various learning algorithms (e.g., \cite{meduri-2020}) often degenerate to random sampling or to sampling from a single class when used with highly stochastic models such as neural networks \cite{beygelzimer-2010, ash-2020}. Furthermore, the computational overhead of training deep neural networks precludes approaches that expect model retraining over many iterations.
 
In this section, we propose an AL sampling strategy facilitated by the decoupling of feature and matching learning tasks and by VAE's latent space. This strategy is characterized by three important properties:
\begin{itemize}[leftmargin=*, nosep]
    \item \textbf{Class balance}: ensures that the samples presented to the user for labeling represent both matches and non--matches and, thus, avoids class imbalance problems.
    \item \textbf{Informativeness}: ensures that the samples presented to the user for labeling maximize the information gain and, thus, potentially speed up the matcher generalization.
    \item \textbf{Diversity}: ensures that the samples selected for user labeling cover a diversified range and, thus, prevents overfitting.
\end{itemize}

Here we emphasize a \textit{third cost--effectiveness property} of our proposal: decoupling the matching model reduces training times to an extent that enables iterative training over multiple active learning iterations using a proposed sampling strategy that eases the manual task of labeling data, as we now describe.

\subsection{Bootstrapping for initial training data}

Contrary to previous deep ER--specific AL approaches (e.g., \cite{kasai-2019}), in this paper we aim to automatically create the initial pool of labeled instances $\mathcal{L}$, factored as two disjoint subsets, $\mathcal{L^+} \cup \mathcal{L^-} = \mathcal{L}$, for matches and non--matches, respectively. To this end, we use Algorithm \ref{alg:al_boot} that relies on the latent space, modeled as described in Section \ref{sec:RL}, to identify the nearest and the furthest entity representations to act as initial positive and negative examples, respectively.

\begin{algorithm}[t]
    \small
	\caption{\small AL bootstrapping}
	\begin{flushleft}
	\textbf{Input}: Tuples $T$, repres. model $\phi$, num. neighbours $k$ \\
 	\textbf{Output}: Pos/Neg/Unlabeled tuple pairs $\mathcal{L}^+$/$\mathcal{L}^-$/$\mathcal{U}$
	\end{flushleft}
	\begin{algorithmic}[1]
	    \Function{ALBootstrap}{}
	   % \State $\mathcal{U} \gets \textproc{Blocking}(T, \phi, k)$
	    \State $\mathcal{U} \gets \{\}$
	    \State $R \gets \phi.\mathsf{predict}(T)$
	    \State $I \gets \mathsf{lsh\_index}(R)$
	    \ForAll{$t_i \in T\ and\ r_i \in R$}
	        \State $N = I.\mathsf{lookup(r_i, k)}$
	        \ForAll{$n_j \in N$}
	            \State $\mathcal{U} \gets \mathcal{U} \cup \{(t_i, n_j)\}$
	        \EndFor
	    \EndFor
	    \State $W_{min} = \min\limits_{(s, t) \in \mathcal{U}} W^2_2(\phi(s), \phi(t))$
	    \State $W_{max} = \max\limits_{(s, t) \in \mathcal{U}} W^2_2(\phi(s), \phi(t))$
	    \State $\mathcal{L}^+ \gets \{(s, t)| (s, t) \in \mathcal{U}, W^2_2(\phi(s), \phi(t)) \approx W_{min}\}$
	    \State $\mathcal{L}^- \gets \{(s, t)| (s, t) \in \mathcal{U}, W^2_2(\phi(s), \phi(t)) \approx W_{max}\}$
	    \State \Return $\mathcal{L}^+, \mathcal{L}^-$, $\mathcal{U} \setminus (\mathcal{L}^+ \cup \mathcal{L}^-)$
	    \EndFunction
	\end{algorithmic}
	\label{alg:al_boot}
\end{algorithm}

Specifically, given a collection of input tuples, $T$, and a representation model $\phi$, trained as described in Section \ref{sec:RL}, we first generate the pool of unlabeled candidates, $\mathcal{U}$, by performing nearest--neighbour search, e.g., using Locality Sensitive Hashing \cite{indyk-1998} with Euclidean distance \cite{datar-2004}, (Lines 3--10), i.e., each candidate is a pair of neighboring tuples $(s, t)$ that may or may not be duplicates. Note that the use of LSH based on Euclidean distance is possible because, looking back to Equation \ref{eq:wasserstein}, we observe that the $W_2$ distance of two $k$--dimensional Gaussian distributions $p=\mathcal{N}(\mu^p, \sigma^p)$ and $q=\mathcal{N}(\mu^q, \sigma^q)$ is \textit{positively correlated} with the squared Euclidean distance of their means, given by $\sum^{k}_{i=1} (\mu^p_i - \mu^q_i)^2$. In other words, duplicate tuples that have $W_2$--close representations in the latent space will have Euclidean--close means as well. This observation allows us to use the Euclidean distance as surrogate for similarity candidates and employ LSH algorithms to efficiently find them. 

Next, we measure the distances between the closest and the furthest two neighbours in $\mathcal{U}$ using Equation \ref{eq:wasserstein} and use them as approximate thresholds for initial positive and negative samples (\textit{Lines 11, 12}). The sets of positive (i.e., $\mathcal{L}^+$) and negative (i.e., $\mathcal{L}^-$) samples are given by pairs with similar distances between their members to the minimum and maximum distances, respectively (\textit{Lines 5, 6}).

The intuition behind Algorithm \ref{alg:al_boot} is that very similar tuples will end up with very small distances between their representations and we can automatically choose them as initial positive samples. Similarly, tuples that are very far apart can provide initial negative samples.

\subsection{Sampling for AL in the latent space}

Having obtained initial $\mathcal{L}^+$, $\mathcal{L}^-$ and $\mathcal{U}$, we now describe how we aim to iteratively improve these sets through a sampling strategy that exhibits the three properties mentioned at the beginning of this section.

\subsubsection{Class balance}

Given the unlabeled pool $\mathcal{U}$, in reality only very few of its instances are positives, while the vast majority are negatives. We therefore need to ensure sampling from both categories. We do so by treating positive and negative candidates separately and discriminate between them using the matching model $\gamma$, trained on $\mathcal{L}^+ \cup \mathcal{L}^-$ as described in Section \ref{sec:Matching}. Specifically, instead of sampling from $\mathcal{U}$, we sample from each $\mathcal{U}^+ = \{(s, t)|(s, t) \in \mathcal{U}, p_{\gamma}(1|(s, t)) > 0.5\}$ and $\mathcal{U}^- = \{(s, t)|(s, t) \in \mathcal{U}, p_{\gamma}(1|(s,t)) \leq 0.5\}$, where $p_{\gamma}(1|(s,t))$ is the probability of a given sample $(s, t)$ to be a positive under the current matching model $\gamma$ for that iteration.

\subsubsection{Informativeness}

Given the unlabeled pool $\mathcal{U}$, AL theory hypothesizes that there is a sub--set in $\mathcal{U}$ of minimal size that offers maximum generalization power to a matching model. We therefore need to get as close as possible to this sub--set in order to maximize the information gain of the resulting model. We do so by using one of the most common measures for the amount of information carried by a potential training sample: the \textit{entropy} of the conditional probability under the model aimed to be improved \cite{burr-2009}. More formally, given $\gamma$ and $\mathcal{U} = \mathcal{U}^+ \cup \mathcal{U}^-$, we can use the entropy measure, given by Equation \ref{eq:entropy} for the binary case, to choose the most informative instances in $\mathcal{U}$ to be labeled by the user.

\begin{equation} \label{eq:entropy}
    \small
    \begin{split}
    H_{\gamma}(s, t) &= -p_{\gamma}(y|(s,t))log(p_{\gamma}(y|(s,t))) \\
    &+ (1 - p_{\gamma}(y|(s,t)))log(1 - p_{\gamma}(y|(s,t)))
    \end{split}
\end{equation}

\noindent where $y$ is the predicted class for a tuple pair $(s, t) \in \mathcal{U}$ and $p_{\gamma}(y|(s,t))$ is the predicted probability of $y$ under the current model $\gamma$. 

Intuitively, $H(s, t)$ is high for pairs of uncertain tuples whose probability of being duplicate/non--duplicate is close to $0.5$, and low otherwise. Therefore, in our sampling strategy, those $(s, t)$ from $\mathcal{U}$ that have a high entropy are most informative to the model.

\subsubsection{Diversity}

Given the unlabeled pool $\mathcal{U}$, in addition to entropy, we also have to consider the distance between the tuple representations of each instance. This is because treating positives and negatives separately is often not enough to ensure class balance, i.e., many of the predicted positives with high entropy are in reality negatives. We therefore sample positive candidates that have a high positive probability and a distance between their tuple representations similar to the distances associated with samples from $\mathcal{L}^+$. However, most of the initial $\mathcal{L}^+$ constituents have very small distances between their tuples. Consequently, the Wasserstein vectors computed in the \textit{Distance} layer of Figure \ref{fig:m_arch} will be similar, if not identical. We therefore need to ensure that the AL sampling strategy chooses more diverse positive candidates. We do so by considering the entire distribution of distances between tuple representations of members of $\mathcal{L}^+$. More specifically, given a set of duplicates $\mathcal{L}^+$ and a representation model $\phi$, we repeatedly sample from each $\phi(s)$ and $\phi(t)$, with $(s, t) \in \mathcal{L}^+$, using the \textit{Sampling} step from Figure \ref{fig:arch}, i.e., VAE's \textit{reparameterization trick} \cite{kingma-2014}, to obtain a distribution of possible Euclidean distances $D^+$ between duplicate representations, as shown in Equation \ref{eq:dist_dist}.

\begin{equation} \label{eq:dist_dist}
    \small
    \begin{split}
	        D^+ &= \{\left\Vert z_i^s - z_i^t \right\Vert_{2} | (s, t) \in \mathcal{L}^+; \\
	            & s_i \in \mathsf{sampling(\phi(s))}; t_i \in \mathsf{sampling(\phi(t))}\}
    \end{split}
\end{equation}

Recall that, given a trained variational encoder $\phi$, every sample from the approximated distribution for a tuple $t$ produced by $\phi(t)$ can act as a viable encoding from which the input can be decoded. Therefore, by repeatedly sampling from each $\phi(s)$ and $\phi(t)$, with $(s, t) \in \mathcal{L}^+$, Equation \ref{eq:dist_dist} produces a distribution of possible distances between duplicate representations\footnote{{\small We need to repeatedly sample, e.g., 1000 samples for each tuple, because $\mathcal{L}^+$ alone often contains insufficient samples to accurately estimate the distance distribution.}}.

Having obtained $D^+$, we can employ Kernel Density Estimators (KDE) \cite{silverman-1986} using Gaussian kernels to estimate a univariate probability density function, $\widehat{f}^+(d)$, over all distances $d \in \mathcal{D^+}$. Then, $\widehat{f}^+(d)$ can be applied on the distances between tuple representations of each unlabeled instance from $\mathcal{U}$ to identify the ones that are most likely to be positives, and use this information in addition to entropy.

\subsubsection{Balanced, informative and diverse sampling}

Algorithm \ref{alg:AL} unifies all the steps discussed in this section. Specifically, once the initial sets of unlabeled/labeled samples have been generated (i.e., \textit{Line 2}), the initial matcher trained (i.e., \textit{Line 3}), and the initial positive probability density function estimated (i.e., \textit{Line 4}), we iteratively proceed as follows:

\begin{algorithm}[t]
    \small
	\caption{\small AL for ER}
	\begin{flushleft}
	\textbf{Input}: Tuples $T$, repres. model $\phi$, num. iterations $I$, num. neighbours $k$\\
 	\textbf{Output}: Matching model $\gamma$
	\end{flushleft}
	\begin{algorithmic}[1]
	    \Function{AL}{}
	    \State $\mathcal{U}, \mathcal{L}^+, \mathcal{L}^- \gets \textproc{ALBootstrap}(T, \phi, k)$
	    \State $\gamma \gets \mathsf{train}(\mathcal{L}^+, \mathcal{L}^-)$
	    \State $\widehat{f}^+(d) \gets \mathsf{KDE}(\mathcal{L}^+)$
 	    \ForAll{$i \in I$}
 	        \State $c^+ \gets (s, t),\ \min\limits_{(s, t) \in \mathcal{U}^+} H_{\gamma}((s, t)) \times \frac{1}{\widehat{f}^+(d((s,t))}$
 	        \State $c^- \gets (s, t),\ \min\limits_{(s, t) \in \mathcal{U}^-} H_{\gamma}((s, t)) \times \widehat{f}^+(d((s,t))$
 	        \State $u^+ \gets (s, t), \ \min\limits_{(s, t) \in \mathcal{U}^+} \frac{1}{H_{\gamma}((s, t))} \times \widehat{f}^+(d((s,t))$
 	        \State $u^- \gets (s, t),\ \min\limits_{(s, t) \in \mathcal{U}^-} \frac{1}{H_{\gamma}((s, t))} \times \frac{1}{\widehat{f}^+(d((s,t))}$
 	        \State $\mathsf{label}(c^+);\ \mathsf{label}(c^-);\ \mathsf{label}(u^+);\ \mathsf{label}(u^-)$
 	        \State $\mathcal{L}^+ \gets \mathcal{L}^+ \cup \{c^+, u^+\};\ \mathcal{L}^- \gets \mathcal{L}^- \cup \{c^-, u^-\}$
 	        \State $\mathcal{U} \gets \mathcal{U} \setminus (\mathcal{L}^+ \cup \mathcal{L}^-)$
 	        \State $\gamma \gets \mathsf{train}(\mathcal{L}^+, \mathcal{L}^-)$
	        \State $\widehat{f}^+(d) \gets \mathsf{KDE}(\mathcal{L}^+)$
 	    \EndFor
 	    \State \Return $\gamma$
	    \EndFunction
	\end{algorithmic}
	\label{alg:AL}
\end{algorithm}

\noindent \textbf{Certain positives}. We identify duplicate candidates with low entropy and high distance likelihoods as certain positives (i.e., \textit{Line 6}). Intuitively, these are instances characterized by a high positive probability under the current model and a distance value between their representations \textit{similar} to the distances associated with $\mathcal{L}^+$.

\noindent \textbf{Certain negatives}. We identify non--duplicate candidates with low entropy and low distance likelihood as certain negatives (i.e., \textit{Line 7}). Intuitively, these are instances characterized by a high negative probability under the current model and a distance value between their representations \textit{dissimilar} to the distances associated with $\mathcal{L}^+$.

\noindent \textbf{Uncertain positives}. We identify duplicate candidates with high entropy and low distance likelihood as uncertain positives (i.e., \textit{Line 8}). Intuitively, these are instances characterized by a low positive probability under the current model (although higher than $0.5$) and a distance value between their representations \textit{dissimilar} to the distances associated with $\mathcal{L}^+$.

\noindent \textbf{Uncertain negatives}. We identify non--duplicate candidates with high entropy and high distance likelihood as uncertain negatives (i.e., \textit{Line 9}). Intuitively, these are instances characterized by a low negative probability under the current model (although higher than $0.5$) and a distance value between their representations \textit{similar} to the distances associated with $\mathcal{L}^+$.

Algorithm \ref{alg:AL} identifies two types of samples for each class, \textit{viz.} certain and uncertain. Uncertain samples have high informative value for the model since they are close to the decision boundary (i.e., high entropy) and have surprising distances between their tuple representations given their predicted class. Conversely, the purpose of certain samples is to prevent overfitting to the selected uncertain instances. 
% In practice, certain samples can often be labeled automatically, i.e., without user input, since they have the highest positive/negative probabilities and the most similar/dissimilar distances to the positive distance distribution. 
Finally, in Algorithm \ref{alg:AL}, user involvement is only required at \textit{Line 10} and, although Algorithm \ref{alg:AL} assumes sampling one instance of each type per iteration, in practice the algorithm can easily be extended to perform batch sampling by choosing the top-$k$ instances at each of \textit{Lines 6,7,8,9}.

\section{Evaluation} \label{sec:Evaluation}

In this paper we set out to decrease the cost associated with ER tasks, and approached this by decoupling feature learning from matching tasks. In this section, we empirically show that this decoupling contributes to the cost reduction desideratum without compromising effectiveness.

\subsection{Experimental setup} \label{subsec:setup}

\subsubsection{Datasets}

We conduct experiments on nine datasets from eight domains. Table \ref{tab:data} shows an overview of the evaluation data used in this section. Each domain (i.e., \textbf{Domain} column) presents two tables (of \textbf{Card.} cardinality) between which we aim to perform ER, with the same \textbf{Arity} and aligned attributes. Each domain also comes with a training set with duplicate and non--duplicate example pairs (of \textbf{Training} size), and a similar, albeit smaller, test set (of \textbf{Test} size). Datasets marked with $^\dag$ are \textit{clean} with few missing values. Datasets marked with $^\ddag$ are \textit{noisy} and more challenging to perform ER on due to their many missing values and unstructured attributes, e.g., product descriptions. Finally, the first seven domains have been previously used in ER benchmarks\footnote{Public datasets, together with their training/test instances, available at \url{www.github.com/anhaidgroup/deepmatcher/blob/master/Datasets.md}} (e.g., \cite{ebraheem-2018, mudgal-2018}), while the last two are private datasets from \textit{Peak AI} with data about clothing products and person contacts.

\begin{table}[t]
\centering
\footnotesize
\caption{\small Datasets used in the experiments}
\begin{tabular}{|c|c|c|c|c|}
\hline
\textbf{Domain}& \textbf{Card.} & \textbf{Arity} & \textbf{Training} & \textbf{Test} \\ \hline
Restaurants$^\dag$ & 533/331 & 6 & 567 & 189 \\ \hline
Citations 1$^\dag$ & 2616/2294 & 4 & 7417 & 2473 \\ \hline
Citations 2$^\dag$ & 2612/64263 & 4 & 17223 & 5742 \\ \hline
Cosmetics$^\ddag$ & 11026/6443 & 3 & 327 & 81 \\ \hline
Software$^\ddag$ & 1363/3226 & 3 & 6874 & 2293 \\ \hline
Music$^\ddag$ & 6907/55923 & 8 & 321 & 109 \\ \hline
Beer$^\ddag$ & 4345/3000 & 4 & 268 & 91 \\ \hline
Stocks$^\ddag$ & 2768/21863 & 8 & 4472 & 1117 \\ \hline
CRM$^\dag$ & 5742/9683 & 12 & 440 & 220 \\ \hline
\end{tabular}
\label{tab:data}
\end{table}

\subsubsection{Baselines and reported measures}

We build our evaluation\footnote{All experiments have been run using \textit{PyTorch} on a \textit{Python 3 Google Compute Engine Backend} with \textit{12 GB RAM} and \textit{GPU acceleration}.} around \textit{Precision (P)}, \textit{Recall (R)} and \textit{F--measure (F1)}, measured on the test datasets. For the purposes of computing these measures, we define a true positive \textit{(tp)}: any pair of tuples marked as duplicate in both the test set and the evaluated results; a false positive \textit{(fp)}: any pair of tuples marked as non--duplicate in the test set and as a duplicate in the evaluated results; and as a false negative \textit{(fn)}: any pair of tuples marked as a duplicate in the test set and as a non--duplicate in the evaluated results. Then, $P=\frac{tp}{tp+fp};\ R=\frac{tp}{tp+fn}; F1=2 * \frac{P \times R}{P + R}$. 

% Intuitively, $P$ measures the ability of the evaluated component not to mislabel true negatives, and $R$ measures the ability of the evaluated component to find all the positive test samples.

For evaluating the representation learning task we consider simple LSH--based top-$k$ nearest neighbour \cite{datar-2004} baselines using \textit{LSA}--, \textit{word2vec (W2V)}--, \textit{BERT}--, and \textit{EmbDI}--generated tuple representations. We compare each of these against a top-$k$ nearest neighbour approach that uses \textit{VAER} representations generated from \textit{LSA}--, \textit{word2vec (W2V)}--, \textit{BERT}--, and \textit{EmbDI}--based $IR$s, respectively. Here, we aim to show the generality of our approach and the effectiveness of the representation learning task over different types of $IR$s.

For evaluating the matching task, we consider \textit{DeepER} \cite{ebraheem-2018}, \textit{DeepMatcher} \cite{mudgal-2018}, and \textit{DITTO} \cite{yuliang-20} as the state--of--the--art supervised ER. These systems, while highly effective, do not allow for transferability or practical training times for reasons mention in Section \ref{sec:Relwork}. We aim to demonstrate \textit{VAER}'s superiority in these respects, while keeping similar levels of effectiveness. Other ER proposals, such as \textit{Magellan} \cite{konda-2016} or \textit{ZeroER} \cite{wu-2020}, are not considered in this paper since they do not rely on deep learning and have already been subject to extensive comparison against this space (e.g., \cite{mudgal-2018}, \cite{wu-2020}).

\subsubsection{VAER configuration}

The most important hyper--parameters have been exemplified in Table \ref{tab:params}. The values have been obtained through hyper--parameter optimization using a validation set that comes with the evaluation datasets. The number of neighbors $K$, and the margin $M$ of the matching loss function are data dependent. However, the values shown in the table led to good results for all the domains used in the evaluation. The configuration of baselines is the default one from their available implementations.

\begin{table}[t]
    \centering
    \footnotesize
    \caption{\small Hyperparameters of VAER}
    \begin{tabular}{|c|c|c|}
    \hline
    \textbf{Component} & \textbf{Parameter} & \textbf{Value} \\ 
    \hline
    \multirow{2}{*}{\begin{tabular}[c]{@{}c@{}}Repr. \\ learning\end{tabular}} & VAE hidden dimension & 200 \\ 
    \cline{2-3} & VAE latent dimension & 100 \\ 
    \hline
    Matching & Margin M & .5 \\ 
    \hline
    AL & Samples/iteration & 10 \\ 
    \hline
    AL & Top neighbours K & 10 \\ 
    \hline
    \multirow{2}{*}{\begin{tabular}[c]{@{}c@{}}Repr. learning \&\\ matching\end{tabular}} & Optimizer & Adam \\ 
    \cline{2-3} & Learning rate & 0.001 \\ 
    \hline
    \end{tabular}
    \normalsize
    \label{tab:params}
\end{table}

\subsection{Representation learning experiments} \label{subsec:rl_exp}

\begin{table*}[t]
\centering
\footnotesize
\caption{\small \textit{VAER} representation learning P/R/F1 showing consistency across all $IR$ types}
\begin{tabular}{|l|p{.2cm}|p{.2cm}|p{.2cm}|p{.2cm}|p{.2cm}|p{.2cm}|p{.2cm}|p{.2cm}|p{.2cm}|p{.2cm}|p{.2cm}|p{.2cm}|p{.2cm}|p{.2cm}|p{.2cm}|p{.2cm}|p{.2cm}|p{.2cm}|p{.2cm}|p{.2cm}|p{.2cm}|p{.2cm}|p{.2cm}|p{.2cm}|}
\hline
\textbf{Domain} & \multicolumn{6}{c|}{\textbf{LSA/VAER$^{LSA}$}} & \multicolumn{6}{c|}{\textbf{W2V/VAER$^{W2V}$}} & \multicolumn{6}{c|}{\textbf{BERT/VAER$^{BERT}$}} & \multicolumn{6}{c|}{\textbf{EmbDI/VAER$^{EmbDI}$}} \\ \hline
& \multicolumn{2}{c|}{\textbf{P}} & \multicolumn{2}{c|}{\textbf{R}} & \multicolumn{2}{c|}{\textbf{F1}} & \multicolumn{2}{c|}{\textbf{P}} & \multicolumn{2}{c|}{\textbf{R}} & \multicolumn{2}{c|}{\textbf{F1}} & \multicolumn{2}{c|}{\textbf{P}} & \multicolumn{2}{c|}{\textbf{R}} & \multicolumn{2}{c|}{\textbf{F1}} & \multicolumn{2}{c|}{\textbf{P}} & \multicolumn{2}{c|}{\textbf{R}} & \multicolumn{2}{c|}{\textbf{F1}} \\ \hline
Rest. &.17&.17&\textbf{1}&\textbf{1}&.29&.29&.31&.23&.95&\textbf{1}&.47&.37&.26&.24&.95&\textbf{1}&.4&.41&.23&.23&\textbf{1}&\textbf{1}&.37&.37 \\ \hline
Cit. 1 &.49&.51&.98&\textbf{1}&.64&.68&.57&.56&.38&\textbf{.98}&.46&.72&.49&.53&.98&\textbf{1}&.65&.69&.5&.47&.89&\textbf{1}&.65&.64 \\ \hline
Cit. 2 &.6&.67&.89&\textbf{.91}&.7&.77&.75&.77&.51&\textbf{.82}&.6&.8&.61&.75&.64&\textbf{.83}&.63&.79&.59&.7&\textbf{.94}&.93&.72&.8 \\ \hline
Cosm. &.65&.68&\textbf{.85}&.83&.74&.76&.74&.65&.84&\textbf{.89}&.78&.76&.65&.78&.7&\textbf{.78}&.67&.78&.66&.75&.14&\textbf{.25}&.24&.35 \\ \hline
Soft. &.21&.25&.72&\textbf{.79}&.33&.39&.22&.23&\textbf{.83}&.8&.35&.36&.26&.29&.6&\textbf{.68}&.37&.41&.28&.28&\textbf{.94}&.93&.43&.43 \\ \hline
Music &.58&.65&.77&\textbf{.82}&.66&.73&.6&.62&.84&\textbf{.85}&.69&.71&.7&.68&.87&\textbf{.93}&.77&.79&.72&.66&.29&\textbf{.86}&.42&.75 \\ \hline
Beer &.44&.48&.84&\textbf{.86}&.58&.62&.44&.5&\textbf{.84}&.8&.58&.62&.47&.57&.78&\textbf{.79}&.59&.67&.7&.64&.91&\textbf{1}&.78&.79 \\ \hline
Stocks &1&1&.79&\textbf{.82}&.88&.9&1&1&.35&\textbf{.45}&.54&.62&1&1&.64&\textbf{.7}&.78&.82&1&.99&.23&\textbf{.77}&.54&.86 \\ \hline
CRM &1&.97&.68&\textbf{.81}&.79&.89&.98&.97&\textbf{.9}&.85&.94&.92&.96&.98&.56&\textbf{.8}&.71&.88&1&.8&\textbf{1}&.88&.1&.84 \\ \hline
\end{tabular}
\label{tab:repr}
\end{table*}

In this subsection, we evaluate the similarity--preserving nature of our entity representations in unsupervised settings. Concretely, we perform LSH top-$K$ nearest--neighbour search, with $K =10$, on $IRs$ generated using each of the techniques from Section \ref{subsec:IR}. We compare the results against a similar nearest--neighbour search on representations generated by the encoding layer of Figure \ref{fig:arch} with inputs of corresponding types. 
% Our hypothesis is that \textit{if tuple representations are similarity--preserving, the representations of duplicates will be close by in the latent space and, therefore, deemed near neighbors.} 
We note that such an approach can also act as a \textit{blocking} step in an end--to--end ER process (similar to \cite{ebraheem-2018}) and, therefore, aim for \textit{high recall} because missed duplicates at this step would be unrecoverable by matching. 

Table \ref{tab:repr} shows the values for precision, recall and F1 score \textit{@ K=10}, for $IR$ nearest--neighbour search\footnote{For each tuple pair in the test set, we measure the effectiveness against the top-$10$ most similar neighbours of either of the two tuples in the pair.} (i.e., left--hand--side values) compared against the results of a \textit{VAER} tuple representations nearest--neighbour search with $IR$ inputs of the corresponding type (i.e., right--hand--side values). Since each representation returned by the variational encoder is a $(\mu, \sigma)$ vector--pair, we perform the search on $\mu$ vectors and reorder the results according to the $W^2_2$ distance from Eq. \ref{eq:wasserstein} to include the $\sigma$ vectors as well.

Overall, the results show the consistency and the potential of VAE encodings to improve the effectiveness of an unsupervised $IR$--based ER task across all types of $IRs$. \textit{LSA} seems to be the most robust $IR$--type choice, with \textit{BERT} and \textit{EmbDI} closely following. In the case of \textit{Cosmetics}, there are many similar entities that only diverge in one attribute, e.g., \textit{color}. The representations of such tuples tend to be very similar, especially when generated using \textit{EmbDI}, which leads to lower recall values. The lowest recall value, i.e., for \textit{Software}, is determined by noisy and missing data. However, \textit{EmbDI} copes well with such data inconsistencies and this behavior is preserved after VAE encoding.

\begin{figure}[t]
\centering
\includegraphics[width=.75\linewidth]{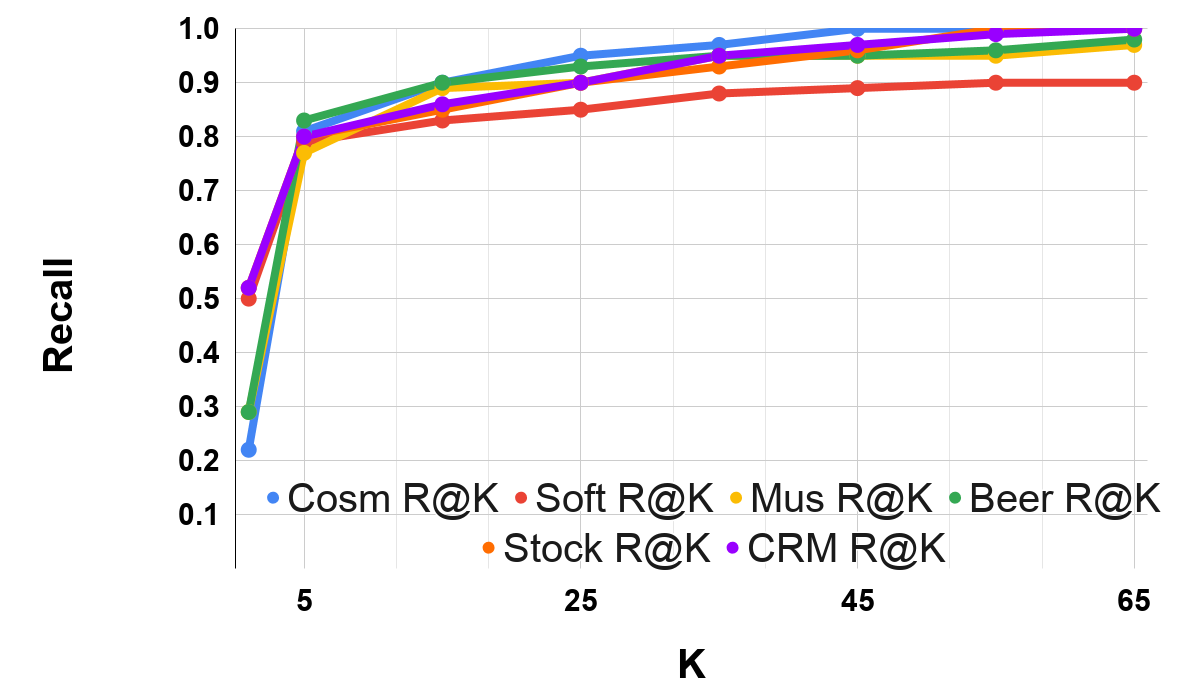}
\caption{\small $VAER^{LSA}$'s recall@$K$ as $K$ increases}
\label{fig:r_k}
\end{figure}

The effectiveness of the nearest--neighbor search depends on the value of $K$. Figure \ref{fig:r_k} shows that most of the $VAER^{LSA}$ cases that did not already achieve good recall in Table \ref{tab:repr} (i.e., the last six domains) can be improved by increasing $K$. 

The experiments from this subsection show that VAE--generated tuple representations are similarity--preserving and that the proposed representation learning method is \textit{cost--effective}, in the sense that it alleviates the user from deciding on the features that govern the similarity between representations, and robust across multiple types of $IR$.

\subsection{Supervised matching experiments} \label{subsec:m_exp}

In this section, we evaluate the supervised matching task, hypothesizing that \textit{our matching model is more efficient than the baselines}, e.g., \textit{DeepER} (DER), \textit{DeepMatcher} (DM), and \textit{DITTO}, \textit{without compromising effectiveness}.

\begin{table*}[t]
\centering
\footnotesize
\caption{\small Similarity learning results showing \textit{VAER}'s matching effectiveness}
\begin{tabular}{|l|c|c|c|c|c|c|c|c|c|c|c|c|}
\hline
\textbf{Domain} & \multicolumn{3}{c|}{\textbf{VAER}$^{LSA}$} & \multicolumn{3}{c|}{\textbf{DER}} & \multicolumn{3}{c|}{\textbf{DM}} & \multicolumn{3}{c|}{\textbf{DITTO}} \\ \hline
& \textbf{P} & \textbf{R} & \textbf{F1} & \textbf{P} & \textbf{R} & \textbf{F1} & \textbf{P} & \textbf{R} & \textbf{F1} & \textbf{P} & \textbf{R} & \textbf{F1} \\ \hline
Rest. &1&.97&\textbf{.99}&.95&1&.97&.95&1&.97&1&.95&.97 \\ \hline
Cit. 1 &.97&1&\textbf{.99}&.96&.99&.97&.96&.99&.97&1&.99&.99 \\ \hline
Cit. 2 &.9&.9&.9&.9&.92&.91&.94&.94&\textbf{.94}&.97&.86&.91 \\ \hline
Cosm. &.87&.94&\textbf{.91}&.83&.96&.89&.89&.92&.9&.91&.81&.86 \\ \hline
Soft. &.62&.64&.63&.62&.62&.62&.59&.64&.62&.72&.71&\textbf{.71} \\ \hline
Music &.86&.86&.86&.78&.9&.83&.95&.81&\textbf{.88}&.78&1&.87 \\ \hline
Beer &.75&.85&.8&.59&.92&.72&.63&.85&.72&.72&.92&\textbf{.81} \\ \hline
Stocks &.99&.99&.99&1&1&\textbf{1}&.99&.99&.99&.99&.98&.98 \\ \hline
CRM &.97&.99&\textbf{.99}&.96&.94&.95&.98&.97&.97&.94&.98&.96 \\ \hline
\end{tabular}
\label{tab:match}
\end{table*}

Table \ref{tab:match} shows the precision, recall and F1 score of the model from Figure \ref{fig:m_arch} using \textit{LSA} $IRs$, \textit{DeepER}, \textit{DeepMatcher}, and \textit{DITTO}, each trained on given training samples for each of the domains in Table \ref{tab:data}. We have empirically chosen \textit{LSA} $IRs$ due to their better performance compared to other $IR$ types. However, the differences are marginal, e.g., on average, the F1--score difference between \textit{LSA} and \textit{W2V}/\textit{BERT} was $0.06$. 

The highlighted values denote cases where \textit{VAER} performs better than the antagonists. However, as with the case when one of the competitors achieves better results, e.g., \textit{Citations 2}, \textit{Stocks}, etc. the differences are minimal. \textit{Software} is a case of particular interest because it proved problematic for all ER solutions evaluated. This is because its tables contain only three columns, one of which is numerical, one contains software product descriptions that are challenging to compare, and the third presents many missing values.

\begin{table}[t]
\centering
\footnotesize
\caption{\small Training times (s)}
\begin{tabular}{|l|c|c|c|c|c|}
\hline
\textbf{Domain} & \multicolumn{2}{c|}{\textbf{VAER}$^{LSA}$} & \textbf{DER} & \textbf{DM} & \textbf{DITTO} \\ \hline
& \textbf{Repr.} & \textbf{Match} &\textbf{Match}&\textbf{Match}&\textbf{Match} \\ \hline
Rest. &\textbf{4.37}&\textbf{2.5}&84.5&258.79&93.51 \\ \hline
Cit. 1 &\textbf{23.5}&\textbf{10.14}&549.65&1022.31&100.94 \\ \hline
Cit. 2 &\textbf{127.84}&\textbf{23.6}&1145.57&2318.89&1523.93 \\ \hline
Cosm. &83.1&1.73&\textbf{33.88}&103.12&84.17 \\ \hline
Soft. &\textbf{21.95}&\textbf{19.43}&552.26&986.07&679.47 \\ \hline
Music &335.32&1.4&\textbf{62.28}&160.15&64.18 \\ \hline
Beer &57.29&4.61&\textbf{33.61}&58.76&59.96 \\ \hline
Stocks &\textbf{182.29}&\textbf{17.29}&836.94&1509.49&436.85 \\ \hline
CRM &81.31&1.88&\textbf{40.23}&121.76&85.83 \\ \hline
\end{tabular}
\label{tab:ttime}
\end{table}

We can conclude from Table \ref{tab:match} that our proposed Siamese matching model \textit{achieves state--of--the--art effectiveness levels}. Additionally, the decoupling of feature learning and matching can \textit{lead to a significant decrease in training time} for some of the evaluated domains, as shown in Table \ref{tab:ttime}. The table compares the combined training times of $VAER^{LSA}$'s representation and matching models\footnote{Using other types of $IRs$ returns marginally similar results.} against \textit{DeepER}, \textit{DeepMatcher}, and \textit{DITTO}. For five out of nine cases highlighted in Table \ref{tab:ttime}, \textit{VAER} requires orders of magnitude lower training times. This is by virtue of $IRs$ that impose reduced input dimensions, and by virtue of the simple architectures of the representation and matching models. The other four cases: \textit{Cosmetics}, \textit{Music}, \textit{Beer} and \textit{CRM} are domains with many tuples to match and reduced amounts of training data. This suggests that \textit{VAER}'s representation training time is dominated by the size of the input tables, while \textit{VAER}'s matching training time, similarly to the baselines, is dominated by the size of the training set. In practice, training representations on large input tables can be accelerated by training on just a sample of all tuples. Alternatively, transfer learning can be employed, as we show in the next experiment.

\subsection{Transferability experiments} \label{subsec:trf_exp}

\begin{table}[t]
\centering
\footnotesize
\caption{\small Recall/F1 score with local/transferred repr. models.}
\begin{tabular}{|l|c|c|c|c|c|c|}
\hline
\textbf{Domain} & \multicolumn{3}{c|}{\textbf{Repr. recall@$K$}} & \multicolumn{3}{c|}{\textbf{Matching F1}} \\ \hline
& \textbf{Local} & \textbf{Transf.} & $\Delta$ & \textbf{Local} & \textbf{Transf.} & $\Delta$ \\ \hline
Rest. &1&1&0&.97&.96&-.01 \\ \hline
Cit. 1 &.99&1&+.01&.99&.97&-.02 \\ \hline
Cit. 2 &.91&.91&0&.9&.9&0 \\ \hline
Cosm. &.83&.83&0&.86&.85&-.01 \\ \hline
Soft. &.8&.79&-.01&.59&.57&-.02 \\ \hline
Music &.79&.75&-.04&.8&.78&-.02 \\ \hline
Beer &.86&.86&0&.79&.77&-.02 \\ \hline
Stocks &.79&.79&0&.95&.97&-.02 \\ \hline
CRM &.81&.84&+.03&.97&.98&+.01 \\ \hline
\end{tabular}
\label{tab:transf_r_f}
\end{table}

\begin{table*}[t]
\centering
\footnotesize
\caption{\small Active Learning results showing data labeling cost--reductions.}
\begin{tabular}{c|c|c|c|c|c|c|c|c|c|cc}
\cline{2-10}
 & \multicolumn{3}{c|}{$\boldsymbol{Precision}$} & \multicolumn{3}{c|}{$\boldsymbol{Recall}$} & \multicolumn{3}{c|}{$\boldsymbol{F1}$} &  &  \\ \hline
\multicolumn{1}{|c|}{\textbf{Domain}} & \textbf{Bootstrap} & \textbf{A250} & \textbf{Full} & \textbf{Bootstrap} & \textbf{A250} & \textbf{Full} & \textbf{Bootstrap} & \textbf{A250} & \textbf{Full} & \multicolumn{1}{c|}{\textbf{F1 \%}} & \multicolumn{1}{c|}{\textbf{Training \%}} \\ \hline
\multicolumn{1}{|c|}{Rest.} & .73 & 1 & .94 & .6 & 1 & 1 & .65 & 1 & .97 & \multicolumn{1}{c|}{\textbf{103\%}} & \multicolumn{1}{c|}{\textbf{44\%}} \\ \hline
\multicolumn{1}{|c|}{Cit. 1} & .96 & .95 & .97 & .84 & .97 & 1 & .89 & .95 & .99 & \multicolumn{1}{c|}{\textbf{96\%}} & \multicolumn{1}{c|}{\textbf{3.3\%}} \\ \hline
\multicolumn{1}{|c|}{Cit. 2} & .9 & .7 & .9 & .33 & .8 & .9 & .48 & .74 & .9 & \multicolumn{1}{c|}{82\%} & \multicolumn{1}{c|}{1.4\%} \\ \hline
\multicolumn{1}{|c|}{Cosm.$^\dag$} & .67 & .8 & .87 & .91 & .85 & .94 & .77 & .82 & .91 & \multicolumn{1}{c|}{\textbf{90\%}} & \multicolumn{1}{c|}{76\%} \\ \hline
\multicolumn{1}{|c|}{Soft.} & .25 & .56 & .62 & .41 & .38 & .64 & .31 & .45 & .63 & \multicolumn{1}{c|}{71\%} & \multicolumn{1}{c|}{3.6\%} \\ \hline
\multicolumn{1}{|c|}{Music} & .46 & .8 & .86 & .63 & .83 & .86 & .53 & .81 & .86 & \multicolumn{1}{c|}{\textbf{94\%}} & \multicolumn{1}{c|}{\textbf{76\%}} \\ \hline
\multicolumn{1}{|c|}{Beer$^\dag$} & .51 & .71 & .75 & .55 & .73 & .85 & .52 & .71 & .8 & \multicolumn{1}{c|}{89\%} & \multicolumn{1}{c|}{92\%} \\ \hline
\multicolumn{1}{|c|}{Stocks$^\dag$} & .99 & .95 & .99 & .83 & .85 & .99 & .90 & .89 & .99 & \multicolumn{1}{c|}{\textbf{90\%}} & \multicolumn{1}{c|}{\textbf{5.5\%}} \\ \hline
\multicolumn{1}{|c|}{CRM} & .83 & .78 & .97 & .63 & .88 & .99 & .71 & .82 & .98 & \multicolumn{1}{c|}{84\%} & \multicolumn{1}{c|}{56\%} \\ \hline
\end{tabular}
\label{tab:al}
\end{table*}

Recall that by decoupling the feature and matching learning tasks, \textit{VAER} enables the transferability of the representation model to other use--cases. In this experiment, we test our hypothesis that \textit{transferability decreases the training--time costs without impacting effectiveness}. We set to prove this hypothesis firstly by observing that, when representation models are transferred from other ER cases, the representation learning times from Table \ref{tab:ttime} are no longer required, making the final training time--cost of \textit{VAER} dependent only on matching model and, therefore, \textit{drastically smaller than the baselines requirements}. Further, in Table \ref{tab:transf_r_f}, we show that the representation learning recall $@K=10$ and the matching F1 scores \textit{remain mostly unchanged} when the representation model is transferred from another domain.

More specifically, we firstly train a \textit{VAER}$^{LSA}$ representation model on all tuples of the \textit{Citations 2} domain (call it the \textit{transferred representation model}). Similarly, we trained seven other \textit{VAER}$^{LSA}$ representation models, one on each of the remaining seven domains (call these the \textit{local representation models}). Then, we measure the recall obtained when performing unsupervised ER (i.e., similar to the experiment from Section \ref{subsec:rl_exp}) on each of the seven domains using the transferred model from \textit{Citations 2} and using the corresponding local representation model. Similarly, we measure the \textit{F1}--score of the matching task when the siamese encoders from Figure \ref{fig:m_arch} are initialized with the parameters of the transferred model and with the parameters of the corresponding local model.

Note that the transferability case restricts the input tables to have the same arity, $a$, expected by the transferred model. Therefore, when the tested dataset has a higher arity we use the first $a$ columns and pad with empty columns datasets with lower arity. This is why the values corresponding to the local models are different from the ones in Tables \ref{tab:repr} and \ref{tab:match}. In practice, a blend of duplicate and non--duplicate tuples from multiple past ER use--cases can be used to create a robust transferable representation model.

Overall, this experiment shows the domain--agnostic nature of the representation learning task, i.e., the representation model can be transferred from other ER tasks with marginal impact on the quality of the representations or the matching effectiveness (see the $\Delta$ columns in Table \ref{tab:transf_r_f}). This property offers a reliable opportunity for using previous representation knowledge and minimizing the ER training time--cost.

\subsection{Active learning experiments}

The practical training times \textit{VAER}'s matching model exhibit in Table \ref{tab:ttime} enable its use in AL schemes where it can be iteratively improved on account of new user--labeled training samples. In this section, we evaluate the cost reduction potential with respect to data labeling of such an AL scheme, i.e., from Section \ref{sec:AL}, hypothesizing that \textit{by actively labeling training samples for the supervised matcher we can still achieve practical effectiveness with less training data.}

Table \ref{tab:al} shows the values for the precision, recall, and F1 scores obtained using three types of matching models: a \textbf{Bootstrap} matching model trained only on data resulting from Algorithm \ref{alg:al_boot}, a \textbf{A250} matching model resulting from Algorithm \ref{alg:AL} with 250 actively labeled samples, and a \textbf{Full} matching model trained on all available training data (from Table \ref{tab:data}). The last two columns denote the percentage of Full model's F1 score the A250 model achieves and the percentage of Full training data size 250 samples represent. Domains marked with $^\dag$ signal cases where the positive samples generated by Algorithm \ref{alg:al_boot} contained false positives that had to be manually removed. On average, Algorithm \ref{alg:al_boot} generated $15$ positives and $15$ negatives (i.e., a balanced initial training set). 

Table \ref{tab:al} highlights examples of cases achieving $90\%$ or more F1 score with less actively labeled samples than provided in the training set. Additionally, cases that have a low Bootstrap precision/recall, e.g., \textit{Software}, \textit{Beer} etc., have low diversity in their bootstrap positive/negative instances. This is expected, since Algorithm \ref{alg:al_boot} retrieves the positives/negatives with the lowest/highest distances between their tuples. Cases that show a significant recall increase from Bootstrap to A250, e.g. \textit{Restaurants}, \textit{Citations 2}, \textit{Beer}, confirm the importance of the diversity property for positive instances. Finally, in only one case, the A250 F1 score percentage is smaller than training size percentage, i.e., \textit{Beer}. Here, $250$ labeled samples achieved $89\%$ of the F1 score obtained with the $269$ samples from the given training set. This is determined by the more diverse positive instances existing in the training data.

\begin{figure}[t]
\centering
\includegraphics[width=.75\linewidth]{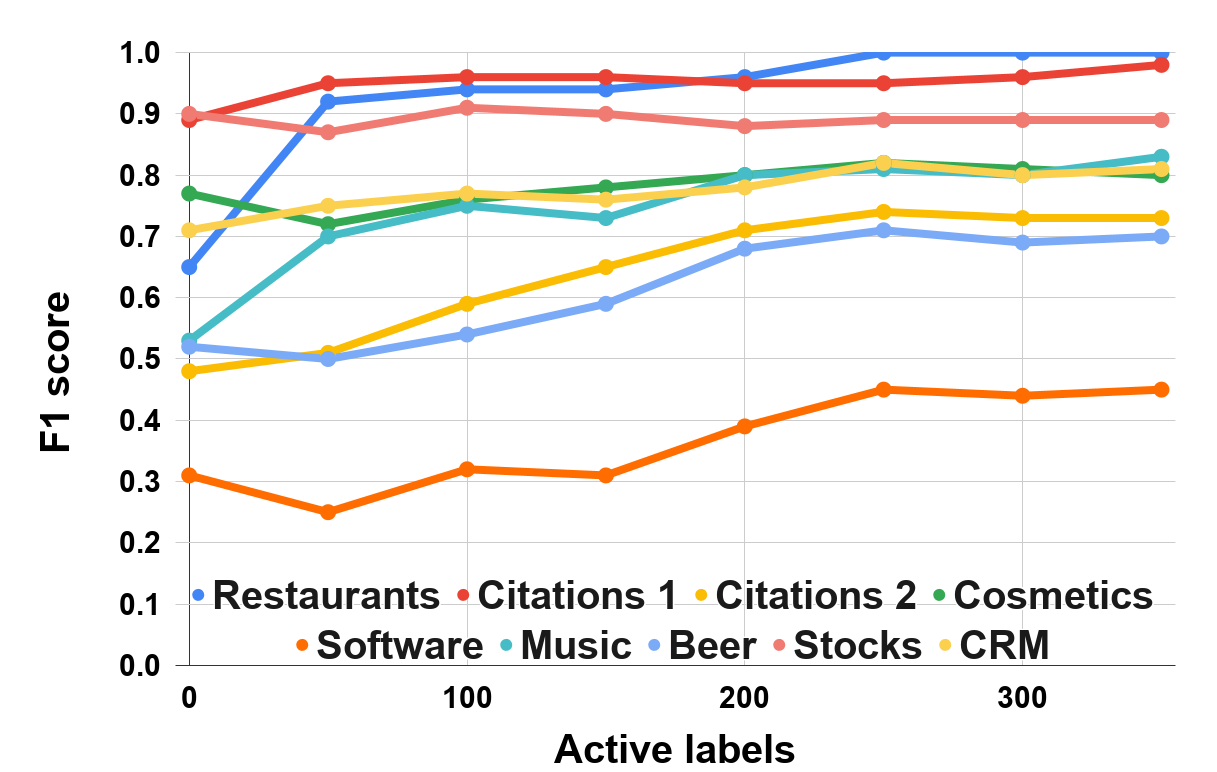}
\caption{\small Active learning F1 score}
\label{fig:al_f1}
\end{figure}

Overall, the last two columns of Table \ref{tab:al} show that our proposed VAE--based AL strategy \textit{leads to reductions in costs associated with data labeling, while achieving practical effectiveness}. However, the effort required to achieve the Full F1 score with actively labeled data is use--case dependent. Consider Figure \ref{fig:al_f1} where the F1 scores obtained with 250 samples remain unchanged even after 100 additional samples across most of the domains. Continuing the AL iterations for the \textit{Citations 2} domain led to achieving $90\%$ of the Full F1 score with 1050 additional actively labeled samples (i.e., for a total of $7.5\%$ of the training set size). Conversely, for the \textit{Software} domain, the same percentage was achieved with just 400 additional samples (i.e., $9.4\%$ of the training set size).

\section{Conclusions}

In this paper we set out to decrease the cost of performing ER with a deep learning model. We identified three requirements associated with ER in practice that generate user--involvement and time costs: (i) the need for features that capture the similarity between duplicates; (ii) the need for duplicate/non--duplicate example data; and (iii) the need to learn a task--specific discriminative similarity function. We approached the cost--reduction desiderata by decoupling the feature and similarity learning tasks. This decoupling, facilitated by with the use of VAEs for the former, allowed us to perform \textit{unsupervised} feature engineering, therefore addressing the cost associated with (i); \textit{fast} supervised matching that \textit{adapts} the unsupervised feature space to the ER case at hand, therefore addressing the cost associated with (iii); and active learning on the matching model, therefore addressing the cost associated with (ii). In addition, we showed how \textit{transferring} a representation model across ER use--cases and data domains can further minimize the cost associated with (i). Lastly, we showed that empirical evaluation supports the fulfillment of our cost--reduction desiderata in practice.

\section{Acknowledgments}
Research supported by a joint collaboration between The University of Manchester, Peak AI Ltd. and Innovate UK as part of a Knowledge Transfer Partnership (KTP11540).

\bibliographystyle{IEEEtran}
\bibliography{main}

\end{document}